# Estimating irregular water demands with physics-informed machine learning to inform leakage detection

Ivo Daniel [a,b,*], Andrea Cominola [a,b]

[a] Chair of Smart Water Networks, Technische Universität Berlin, Straße des 17. Juni 135, 10623 Berlin, Germany

[b] Einstein Center Digital Future, Wilhelmstraße 67, 10117 Berlin, Germany

[*] Email: ivo.daniel@tu-berlin.de

**ABSTRACT**

Leakages in drinking water distribution networks pose significant challenges to water utilities, leading to infrastructure failure, operational disruptions, environmental hazards, property damage, and economic losses. The timely identification and accurate localisation of such leakages is paramount for utilities to mitigate these unwanted effects. However, implementation of algorithms for leakage detection is limited in practice by requirements of either hydraulic models or large amounts of training data. Physics-informed machine learning can utilise hydraulic information thereby circumventing both limitations. In this work, we present a physics-informed machine learning algorithm that analyses pressure data and therefrom estimates unknown irregular water demands via a fully connected neural network, ultimately leveraging the Bernoulli equation and effectively linearising the leakage detection problem. Our algorithm is tested on data from the L-Town benchmark network, and results indicate a good capability for estimating most irregular demands, with $R^2$ larger than 0.8. Identification results for leakages under the presence of irregular demands could be improved by a factor of 5.3 for abrupt leaks and a factor of 3.0 for incipient leaks when compared the results disregarding irregular demands.

## 1 Introduction

Leakages are the most common cause of infrastructure failure in drinking water distribution networks (WDN) (Arregui et al., 2018) and entail a variety of unintended consequences, ranging from operational disruptions (Misiunas et al., 2006), environmental hazards (Wan et al., 2022), property damage (Mansour-Rezaei and Naser, 2013), and sanitary issues (Shortridge and Guikema, 2014) to increased economic cost through non-revenue water (NRW) and infrastructure maintenance (Puust et al., 2010). In 2019, the World Bank estimated global water losses to about 120 million m$^3$ p.a., resulting in an associated economic loss of 39 billion USD p.a. (Liemberger and Wyatt, 2019).

While up to 35% of water supplied to the WDN is lost as NRW in developed countries (Levinas et al., 2021), water losses amount to more than 50% in developing countries (Puust et al., 2010). Additionally, urbanisation (Flörke et al., 2018) and climate change induced drought increase (Konapala et al., 2020) further exacerbate challenges faced by water utilities regarding the sustainability and security of their drinking water provision. Furthermore, leakages add up to an already heavy energy balance of the water sector, which contributes up to 2.7% of total global energy use (Liu et al., 2016), thus requiring drastic climate mitigation solutions (Parkinson, 2021).

In light of these challenges, leakage management arises as a crucial task for water utilities to efficiently sustain and maintain their WDN (Barton et al., 2019). As possible strategies within leakage management, *leakage control* focuses on the implementation of policies and management strategies to prevent and mitigate leakage impacts, while *leakage detection* aims at timely identifying and precisely locating occurring leaks (Puust et al., 2010). Methods for leakage detection can be further divided into (i) *model-based* approaches that employ hydraulic models to compare simulated to sensor data, (ii) *data-driven* approaches that analyse time-series data using mathematical models, and (ii) *hybrid* approaches combining elements from both former categories (Zaman et al., 2020).

In recent years, the leakage detection problem has received increased attention from the scientific community with a steadily increasing body of literature (Romero-Ben et al., 2023), culminating in the global Battle of the Leakage Detection and Isolation Methods (BattLeDIM) (Vrachimis et al., 2022). However, methods for leakage detection applied in practice by water utilities still mostly only consist of periodic water audits combined with the application of in-situ acoustic devices, altogether not sufficing for continuous monitoring (Zaman et al., 2020). This gap between practical application and scientific advances could be caused by method-specific requirements and practical hurdles faced by utilities during a leakage detection method's implementation process. For instance, while model-based and hybrid approaches are capable of providing accurate identification and localisation results, they however depend on the availability of a well-calibrated hydraulic model (Rajabi et al., 2023; Steffelbauer et al., 2022). The quality of the detection results is thus significantly affected by the uncertainty of the input data and model parameters, e.g., attributes of physical components, geographic information, and most importantly water demand data, ultimately limiting the application of model-based approaches in practice (Wan et al., 2022).

Conversely, data-driven algorithms may only require a set of leak-free data for algorithm calibration and are furthermore capable of efficiently analysing high frequency data streamed from online sensors (Romero-Ben et al., 2023). While early studies primarily employ statistical analysis (Eliades and Polycarpou, 2012; Lee et al., 2005; Palau et al., 2012) and probabilistic frameworks (Hutton and Kapelan, 2015; Soldevila et al., 2016), recent advances in machine learning (ML) have accelerated the development of purely data-driven methods (Fu et al., 2022; Kammoun et al., 2022). For instance, convolutional neural networks (CNN) can facilitate accurate leakage identification by analysing pressure data, utilising two-dimensional spectrograms (Guo et al., 2021) or one-dimensional time-series data (Tornyeviadzi and Seidu, 2023), while deep learning was successfully applied on pressure data for precise burst localisation in a small WDN (Zhou, 2019). However, data-driven algorithms require large amounts of training data and an increasing computational capacity, while not yet achieving the detection accuracy of model-based approaches (Vrachimis et al., 2022).

On the other hand, available physical information about hydraulic networks greatly remains disregarded in data-driven leakage detection algorithms, although its inclusion has proven beneficial in many civil engineering domains including applications in urban water management (Vadyala et al., 2022). For instance, physically informing a surrogate model for urban drainage simulations significantly improves its performance with regards to accuracy and training efficiency (Palmitessa et al., 2022).

Here, we present a purely data-driven, pressure-based, and physics-informed leakage detection algorithm. This algorithm builds on previous work that already incorporates hydraulic information in data-driven leakage detection (Daniel et al., 2022) by adding an enhanced capability to deal with unknown irregular water demands (e.g., due to industrial water demand) by employing physics-informed ML. This feature substantially increases the robustness of leakage identification against false negatives, i.e., missed detections.

In detail, physics-informed ML is here employed in the form of an inductive bias (Karniadakis et al., 2021) where physical knowledge about WDN hydraulics is hard-coded into a layer of the ML module. The advantage gained thereby is twofold. Firstly, the explicit estimation of previously unknown irregular demands enables their extraction from the pressure data resulting in additional knowledge about their temporal trajectories. Secondly, feeding the estimations back into the leakage identification algorithm yields additional robustness as these modelled patterns are removed from the data analysed with subsequent change point detection (CPD), resulting in less noise and, thus, more reliable and faster identification of the leakages.

The rest of the paper is organised as follows. In Section 2 we present the mathematical foundation of our leakage identification algorithm and outline the experimental settings we considered in this study. We then present the numerical results to assess the leakage identification performance, including an uncertainty quantification due to the non-deterministic training of our leakage detection algorithm and a sensitivity analysis to varying CPD hyperparameters in Section 3 and discuss them in Section 4. Finally, we summarize final remarks and further research avenues in the last section.

## 2 Material and methods

### 2.1 Mathematical modelling for leakage identification

The foundation of our pressure-based algorithm for data-driven leakage identification is based in the Bernoulli principle, as described in detail in Daniel et al. (2022). Therefrom, the relationship between the pressure $P$ at each pair of sensors $i, j \in N$ can be described via the following linear function:

$$k_i^0 + k_i^1 P_i + \sum_{d \in D} k_i^d Q_d^2 = k_j^0 + k_j^1 P_j + \sum_{d \in D} k_j^d Q_d^2 \qquad (1)$$

where $P$ is the pressure, $Q_d$ represents an irregular flow pattern within a set of irregular water demands $D$ ($d \in D$), possibly occurring anywhere in the WDN (not necessarily at sensor locations $i$ or $j$), and $k^0$, $k^1$, and $k^d$ are coefficients that can be estimated via ordinary least squares (OLS) linear regression when historical time series of $P$ and all $Q_d$ are available. The set of irregular water demands $D$ can be further partitioned into a subset of known irregular demands $D_k$ for which measurements are available and another subset of unknown irregular demands $D_u$.

To construct a model that accounts for all sensors in the WDN simultaneously, each possible combination of $i$ and $j$ in Eq. (1) must be regarded. Accordingly, we expand both sides of Eq. (1) from the single sensor combination $(i, j)$ to $N$ dimensions, resulting it in the following vectorised formulation:

$$\left(k^0 + k^1 P + \sum_{d \in D} k^d Q_d^2\right) \otimes 1_N = 1_N \otimes \left(k^0 + k^1 P + \sum_{d \in D} k^d Q_d^2\right) \qquad (2)$$

where $k^0, k^1, k^d \in \mathbb{R}^N$ represent the combined coefficients from Eq. (1) for $i, j \in [1, ..., N]$, $1_N \in \mathbb{R}^N$ is an expansion vector with all ones, and $1_N \otimes k$ returns the outer product of $1_N$ and $k$. We then solve for $P$ on the left-hand side to yield the combined linear regression equation in $N \times N$ dimensions:

$$\hat{P} = P \otimes 1_N = \frac{1_N \otimes k^0 - k^0 \otimes 1_N}{k^1 \otimes 1_N} + \frac{1_N \otimes k^1}{k^1 \otimes 1_N}(1_N \otimes P) + \sum_{d \in D} \frac{1_N \otimes k^d - k^d \otimes 1_N}{k^1 \otimes 1_N} Q_d^2 \quad (3)$$

where $\hat{P}$ represents the pressure estimated through linear regression. Note that $\hat{P}$ contains $N$ estimations for $P$, one in each column. As a result of the formulation of the $N \times N$ linear regression problem in Eq. (3), the number of regression coefficients could be reduced from $3N^2$ to $3N$ compared to the original formulation in Eq. (1).

Starting from the generalized linear formulation in Eq. (3), leakage identification is facilitated by calculating the model reconstruction error $MRE$ as the difference between the observed pressure values $P_{obs}$ and the estimated pressure values $\hat{P}$:

$$MRE = P_{obs} \otimes 1_N - \hat{P} \quad (4)$$

We then analyse the $MRE$ by means of the *cusum* algorithm for change point detection (Montgomery, 2020) to determine whether significant changes in its distribution over time occur that may represent possible leakages. The *cusum* algorithm detects change points by summing up either the positive or negative deviation of the $MRE$ from its mean value under consideration of a slack value $\delta$. A leak alarm is thereby raised when the *cusum* statistic exceeds a predefined threshold $\varepsilon$.

## 2.2 Physics-informed machine learning for irregular water demand modelling

The linear relationship between the pressure at two sensor locations in a WDN as expressed in Eq. (1) is the result of mostly regular water demands that follow a diurnal pattern, such as the example displayed in Figure 1a, thus, leading to a spatial correlation of the pressure losses. Due to the formulation in Eq. (3), disregarded irregular demands (Figure 1b) directly translate into the $MRE$, hence, possibly obscuring any concurrent leak signal due to their long-tailed probability distribution (Figure 1d) that significantly differs from the distribution of diurnal demand patterns illustrated in Figure 1c. Unmeasured industrial demands of irregular nature substantially affect the MRE and hamper the leakage identification process as noticeable by a comparison between Figure 1e and f. While some irregular demands, such as pump flows, may be known from sensor data ($d \in D_k$) and, thus, can be readily incorporated into the linear regression model, the irregular demands that remain unknown ($d \in D_u$) are not explicitly accessible.

Consequently, the only viable approach to incorporate these unknown irregular demands into the leakage identification process involves extracting them from the available pressure data and simultaneously reintegrating them into the linear regression model. To this end, we employ a physics-informed ML model based on a fully connected neural network (FCNN) for the estimation of the irregular demands that are implicitly contained in the pressure information collected from a WDN.

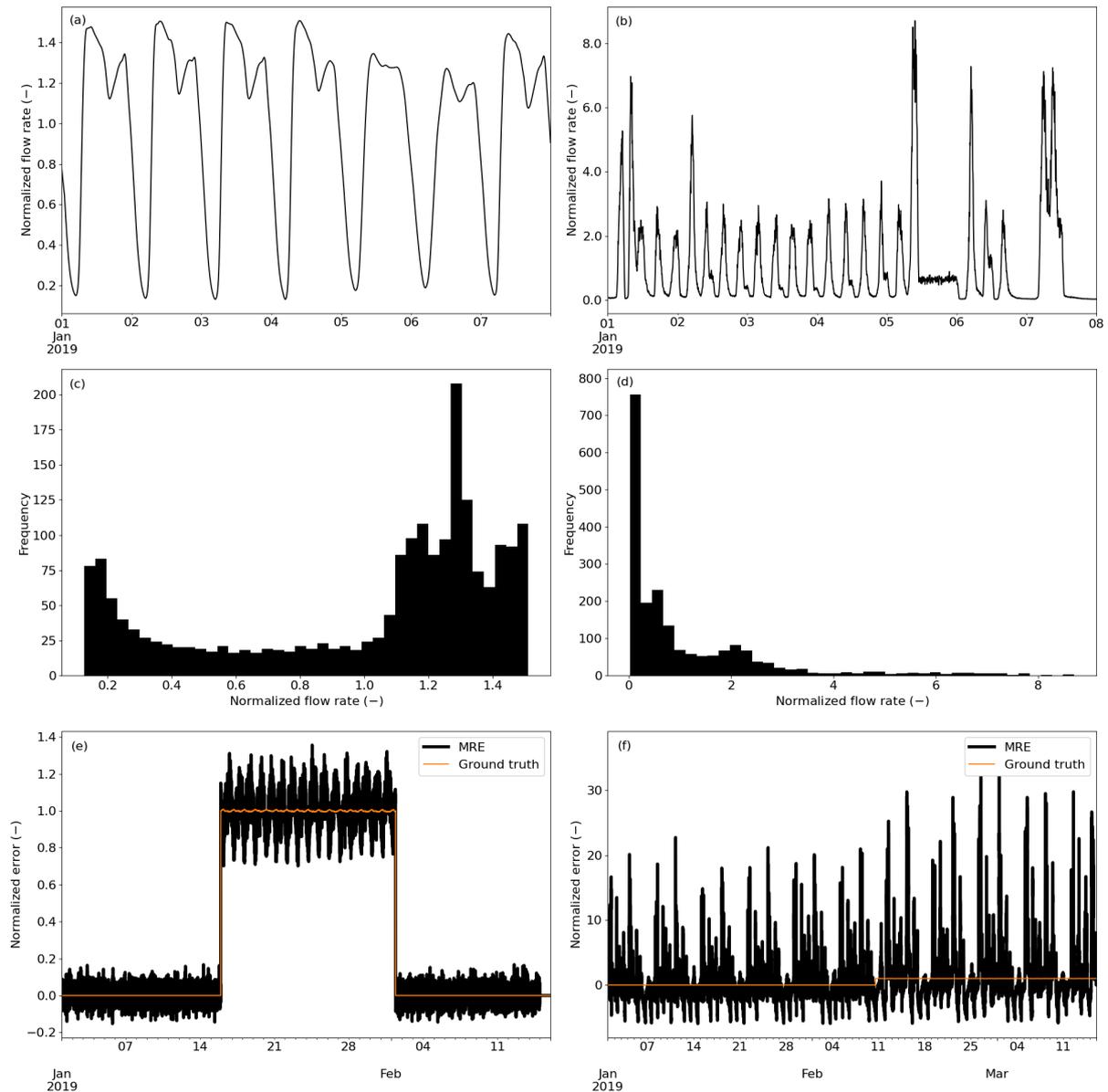

**Figure 1: Characteristics of diurnal vs. irregular demand patterns and their effect on the performance of the leakage identification task.** (a) diurnal demand patterns, (c) their frequency distribution, and (e) their resulting MRE vs. (b) irregular demand patterns, (d) their frequency distribution, and (f) their resulting MRE.

In general, a FCNN, denoted as $\mathcal{N}: x \to y$, represents a mathematical function mapping the input $x$ to the output $y$ by performing a series of matrix operations and applying non-linear activation functions (Goodfellow et al., 2016). A FCNN may be formulated as follows:

$$y = \mathcal{N}(x, \lambda) = \alpha_{L+1} \circ \ldots \circ A_0 \circ \alpha_0(x) \tag{5}$$

where $\lambda$ represents the set of hyperparameters, including the number of hidden layers $l \in [0, \ldots, L]$ and the size of each layer $S$ (i.e., number of neurons). $\alpha_l$ denotes the affine transformation function of layer $l$, and $A_l$ the activation function of layer $l$. The affine transformation function $\alpha_l$ is expressed as:

$$\alpha_l(\xi_{l-1}) = W_l \xi_{l-1} + b_l \tag{6}$$

where $\xi_{l-1}$ is the output of the previous layer $l - 1$ with $\xi_0 = x$, $W_l$ are the weights of layer $l$, and $b_l$ represents the bias of layer $l$. The activation function $A_l$ in Eq. (5) scales the output of the respective layer $l$ before passing it to the subsequent layer (Ramsundar and Zadeh, 2018). For the task of estimating irregular demands from pressure data in Eq. (7), we consider that the input $x = P$ and the output $y = Q_d$.

$$Q_d = \mathcal{N}(P, \lambda) = \alpha_{L+1} \circ \ldots \circ A_0 \circ \alpha_0(P) \tag{7}$$

Since there are no directly observable data available for the latent irregular demands $Q_{d \in D_u}$, which are merely implicitly contained in the observable pressure data, the training process of the FCNN cannot rely on a conventional observational bias. Instead, within our framework, the output of the FCNN is directly fed back into the linear regression model Eq. (3), and the error is quantified as the mean squared error (MSE) on the $MRE$ as calculated in Eq. (4). Consequently, the FCNN is trained implicitly by establishing a feedback loop with the linear regression layer. As the physical information of the hydraulic network is hardcoded into this regression layer, this approach introduces an inductive bias to guide the training of the FCNN in a physics-informed manner (Karniadakis et al., 2021).

Furthermore, we exploit that demands, intended as a physical measure of water abstraction from the WDN, can only obtain non-negative values. Therefore, we introduce an additional rectified linear unit (ReLU) activation function $A_{L+1}^R$ that is applied after the output layer, thus, ensuring $Q_d \geq 0$ (Maas et al., 2013). This modification of Eq. (7) is reflected in Eq. (8).

$$Q_d = \mathcal{N}(P, \lambda) = A_{L+1}^R \circ \alpha_{L+1} \circ \ldots \circ A_0 \circ \alpha_0(P) \tag{8}$$

Finally, batch normalisation is applied to facilitate a stable training procedure, given the implicit nature of the FCNN output data (Santurkar et al., 2018), and k-fold cross-validation (k = 5) is performed to further handle potential noise present in the sensor data (Brunton and Kutz, 2019). We implemented the entire model architecture, including the leakage detection algorithm, using the PyTorch machine learning framework (Paszke et al., 2019).

## 2.3 Case study

We test our physics-informed leakage detection algorithm on the benchmark dataset based on the L-town WDN first developed and released for the Battle of the Leakage Detection and Isolation Methods (BattLeDIM) (Vrachimis et al., 2022). The L-town WDN is composed of three district metered areas (DMA), consisting in total of 785 nodes, 905 pipes, one pump, three valves, and two reservoirs. Moreover, a total of 33 pressure sensors are distributed throughout the network. A schematic of the network layout including highlighted sensor locations is displayed in Figure 2.

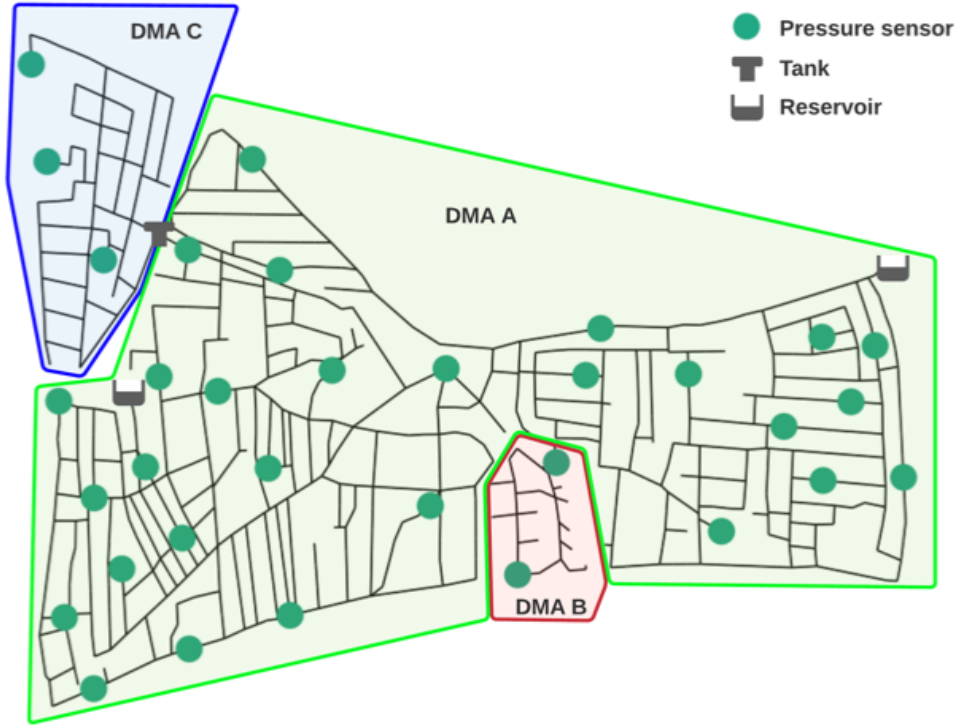

**Figure 2: Schematic of the L-town water distribution network**. Pressure sensors are represented by the green markers. (Source: Daniel et al., 2022)

Data from the pressure sensors are available at a five-minute sampling interval for a time span of two years from the start of 2018 until the end of 2019. Over these two years, a total of 33 leaks were inserted into the WDN, employing the disaster scenario analysis toolbox of the WNTR simulator (Klise et al., 2017).

Start dates for leakages that were imposed on the network are only accessible for the year 2019. Moreover, for the evaluation of our algorithm, only data from the DMA C is relevant as it is the only DMA where unknown irregular demands are present and hamper the leakage identification process. Essentially, for this study we consider the data from the original BattLeDIM dataset for DMA C during 2019, where two different leakages were imposed on the WDN (Vrachimis et al., 2020). The first leak is an abrupt leak or burst which reaches its maximum leakage flow of 5.2 m$^3$/h instantaneously as the result of a pipe break, starting on February 10$^{th}$, 2019, at 13:05. The second leak is an incipient leak with gradually increasing flow rate, e.g., due to a crack evolving over time, starting on May 30$^{th}$, 2019, at 21:55 and reaching its maximum flowrate of 7.2 m$^3$/h around August 11$^{th}$, 2019 (Vrachimis et al., 2020). As any anomaly detection algorithm, our leakage identification algorithm needs to be trained on data that does not contain any anomalies. To this end, we manually selected two weeks of data without leakages for the purpose of training the algorithm. Data from January 1$^{st}$ until 14$^{th}$, 2019, were used as training set for the detection of the first leak. The algorithm was then retrained with data from April 1$^{st}$ until 14$^{th}$, 2019, for the detection of the second leak,

## 2.4 Algorithm evaluation

In this study, we benchmark our physics-informed ML model presented in sections 2.1-2.2, further referred to as LILA-PINN, against the baseline LILA model presented in Daniel et al. (2022), which does not account for the presence of irregular water demands. This model is further referred to as LILA-BASE. Furthermore, we introduce a regression model for additional comparison for which we assume full, perfect knowledge of the industrial (irregular) demands, hereafter referred to as LILA-FK. The latter, however, represents an ideal situation merely presented as an upper limit for comparison.

For a complete evaluation of our algorithm, we further perform an uncertainty quantification due to the non-deterministic training of our leakage detection algorithm and a sensitivity analysis to varying CPD *cusum* hyperparameters, both elaborated on in detail hereafter.

The irregular demands in DMA C are only implicitly observable in an aggregated form. Therefore, their extraction becomes underdefined resulting in non-deterministic training performance of LILA-PINN. We account for this uncertainty by evaluating the performance of LILA-PINN over 100 independent training runs. Its leakage identification capability is quantified with three performance metrics, namely *Recall*, *Precision*, and *F1-score* (Lever et al., 2016), for which formulations are given in Eq. (9) – (11). We adapt the formulation of these metrics to reflect how many model runs have correctly identified each leak, i.e., *true positives* (TP) count, how many missed the leak, i.e., *false negatives* (FN) count, or raised an alarm too early, i.e., *false positives* (FP).

$$Recall = \frac{TP}{TP + FN} \tag{9}$$

$$Precision = \frac{TP}{TP + FP} \tag{10}$$

$$F1 = \frac{2 \cdot TP}{2 \cdot TP + FN + FP} \tag{11}$$

We also consider the time to detection (TTD) of a leak as the difference between its time of detection $t_d$ and its starting time $t_{start}$, as given in Eq. (12) (Taormina et al., 2018).

$$TTD = t_d - t_{start} \tag{12}$$

Furthermore, the leakage identification performance of LILA-PINN greatly depends on the performance of the *cusum* algorithm. Hence, we analyse its sensitivity to the *cusum* hyperparameters. We subject all 100 LILA-PINN models to this sensitivity analysis while investigating the following hyperparameter ranges for the slack value and the threshold value: $\delta = [0, 0.25, ..., 2]$ and $\varepsilon = [200, 225, ..., 400]$.

## 3 Results

In this section, we first show the capability of the LILA-PINN model to extract the irregular water demands from the time series of observed pressure data. Second, we benchmark the leakage identification performance of LILA-PINN against the baseline model, i.e., LILA-BASE, and the model under ideal conditions with perfect knowledge of the irregular demands, i.e., LILA-FK. Finally, we display the results of the uncertainty and sensitivity analyses.

### 3.1 Extraction of irregular water demands

As part of LILA-PINN, the FCNN shall learn to estimate the irregular demands that are implicitly contained in the observable pressure data. To understand the capability of the FCNN for this prediction task, we retrieve the ground truth data of the industrial demands (i.e., irregular demands) from the hydraulic model provided in Vrachimis et al. (2020) and compare it to the estimates provided by the FCNN. Figure 3 illustrates this comparison for the three industrial demands contained in DMA C of the BattLeDIM dataset.

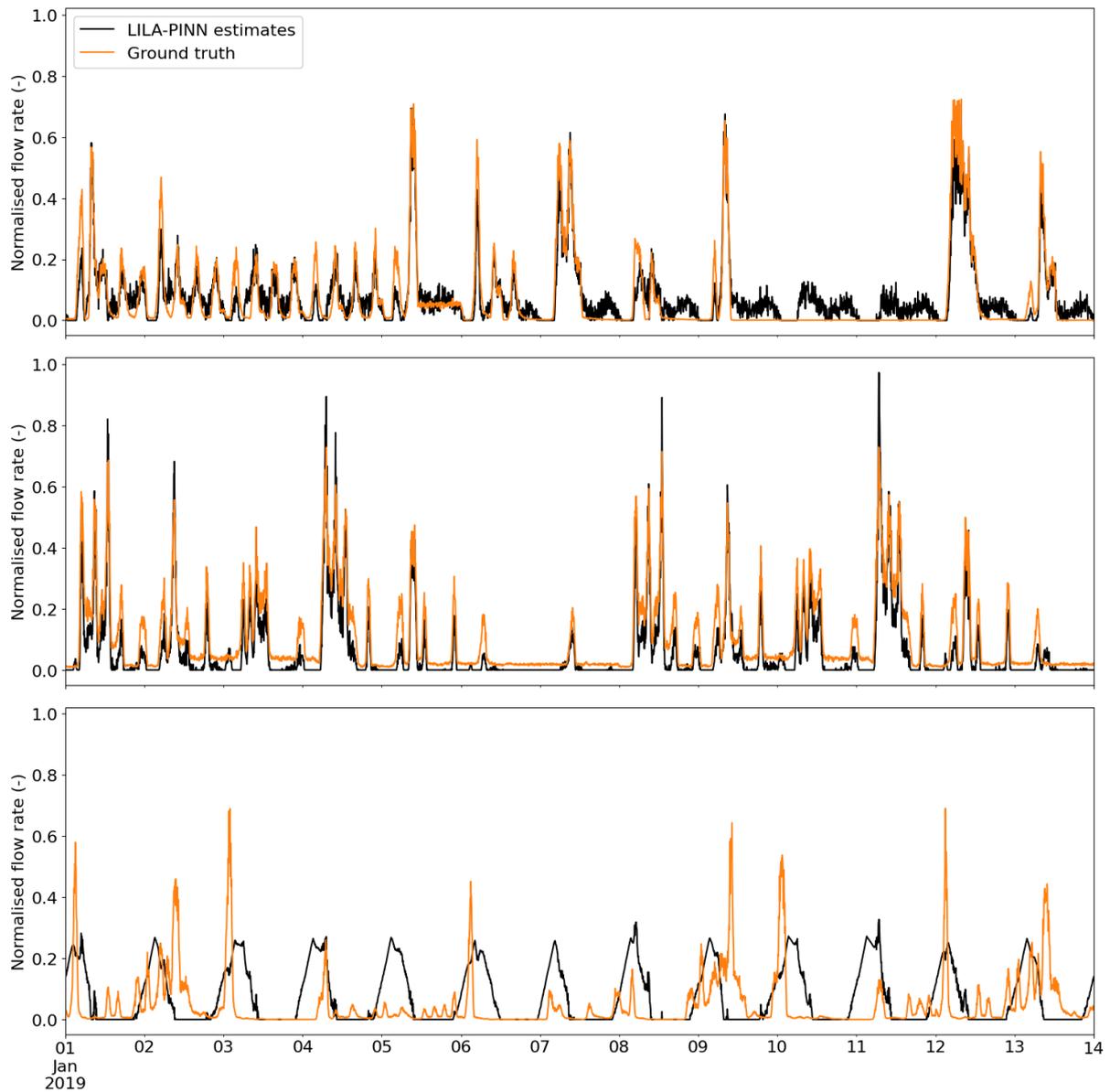

**Figure 3: Comparison of estimated industrial demand patterns to ground truth data.** Each subplot (a) - (c) represents one of three industrial demand patterns present in DMA C of L-town. The flowrates are normalised using a Maximum-Absolute scaler.

The results in Figure 3 demonstrate an overall good capability of the FCNN to predict two out of the three the industrial demands. These demand patterns are nicely replicated, especially with regards to their temporal fluctuations and the relative magnitude of the peaks. The model estimates displayed in Figure 3a and b achieve R2-values of 0.83 and 0.84, and root mean squared error (RMSE) values of 0.058 and 0.057, respectively. This allows for drawing conclusions about unknown irregular demand patterns and could for instance further enable improvement of hydraulic models when applied to real world pressure sensor data.

On the other hand, the FCNN fails to predict one out of the three industrial demands completely (Figure 3c). However, when inspecting this third prediction we observe periodic residuals of the diurnal water demand patterns that are consistent in DMA C. Apparently, in the case of DMA C, the linearity assumption in Eq. (1) resulting from the spatial correlation of all diurnal patterns cannot be perfectly fulfilled. Hence, some residual of the diurnal demand patterns remains with a larger magnitude than the third industrial demand pattern and is thus discovered by the FCNN.

## 3.2 Leakage identification performance

The primary aim of any version of LILA is to promptly identify leakages, aiming for minimal $TTD$, while simultaneously upholding a high level of detection accuracy and minimizing false alarms. LILA-BASE has already demonstrated competitive performance when compared to contemporary state-of-the-art leakage detection algorithms based on economic scores formulated in the BattLeDIM (Daniel et al., 2022). With LILA-PINN, we expand upon LILA-BASE by incorporating a prediction model for the irregular demands only present in DMA C of the L-town WDN. We demonstrate the ability to mask these demands, thereby further enhancing the efficacy of the leakage identification task. The results of this comparison obtained from a single model evaluation are presented in Table 1. The model used for this comparison correspond to the median $TTD$ established during the uncertainty quantification (see Section 3.3).

Table 1: Comparison of time to detection (TTD) for different algorithm versions and leak characteristics. $\delta = 1$ and $\varepsilon = 300$

| Version | Abrupt leak (h) | Incipient leak (d) |
| --- | --- | --- |
| LILA-PINN | 21 | 28.2 |
| LILA-BASE | 119 | 35.6 |
| LILA-FK | 25 | 17.0 |

The results in Table 1 demonstrate the improved performance of LILA-PINN when compared to LILA-BASE. The $TTD$ in case of the abrupt leak is reduced by 82% from 119 hours to 21 hours and in case of the incipient leak by 21%, from 35.6 days to 28.2 days. Notably, in the case of abrupt leaks, LILA-PINN even outperforms the ideal scenario by an additional 16% reduction in $TTD$. This may potentially be attributed to the discovery of the residual patterns originating from the diurnal demand.

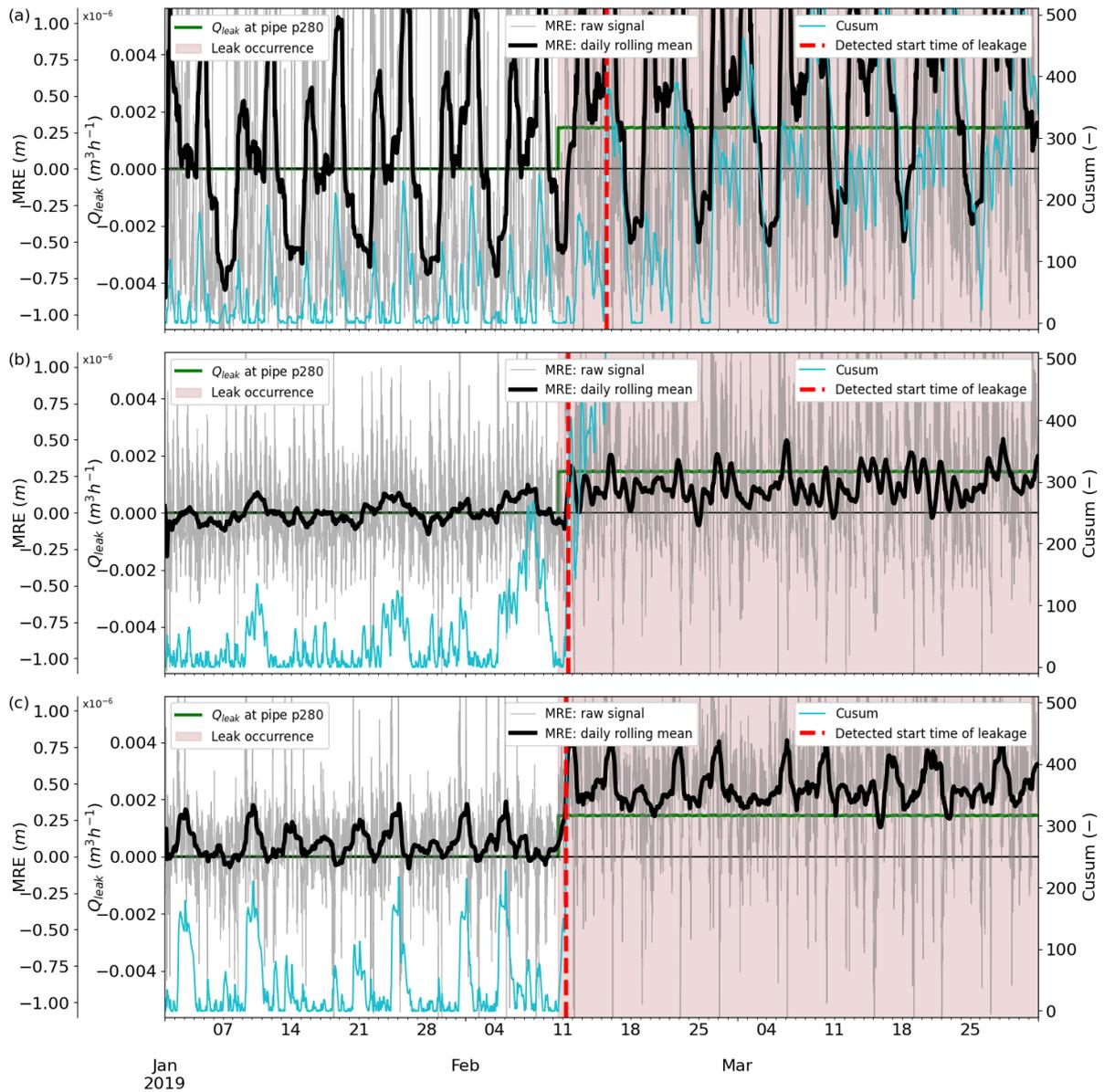

**Figure 4: Comparison of detection performance for the abrupt leak (no. 1) for different algorithm versions: (a) LILA-BASE, (b) LILA-FK, (c) LILA-PINN.** MRE is plotted against the ground truth leak flow for an abrupt leak starting on 2019-02-10 13:05. The Cusum test statistic evaluated for the MRE raw signal is displayed on an additional axis on the right. *TTD* was calculated for $\varepsilon = 300$.

This discovery seems to further enable the leakage identification performance under presence of an abrupt leak. This can be observed in Figure 4c, where the *MRE* resulting from LILA-PINN appears to be ridded of most of the recurring residuals present in Figure 4a of LILA-BASE. Moreover, the MRE in Figure 4c exhibits an even clearer mean shift after the occurrence of the leak when compared to LILA-FK (Figure 4b), leading to an even lower *TTD*.

While the suppression of the recurring residuals with LILA-PINN can also be observed for the incipient leak in Figure 5, a difference in mean shift compared to the ideal scenario with LILA-FK is not apparent. Hence, *TTD* of LILA-PINN for the incipient leak ranks in between the baseline and the ideal scenario (see Table 1).

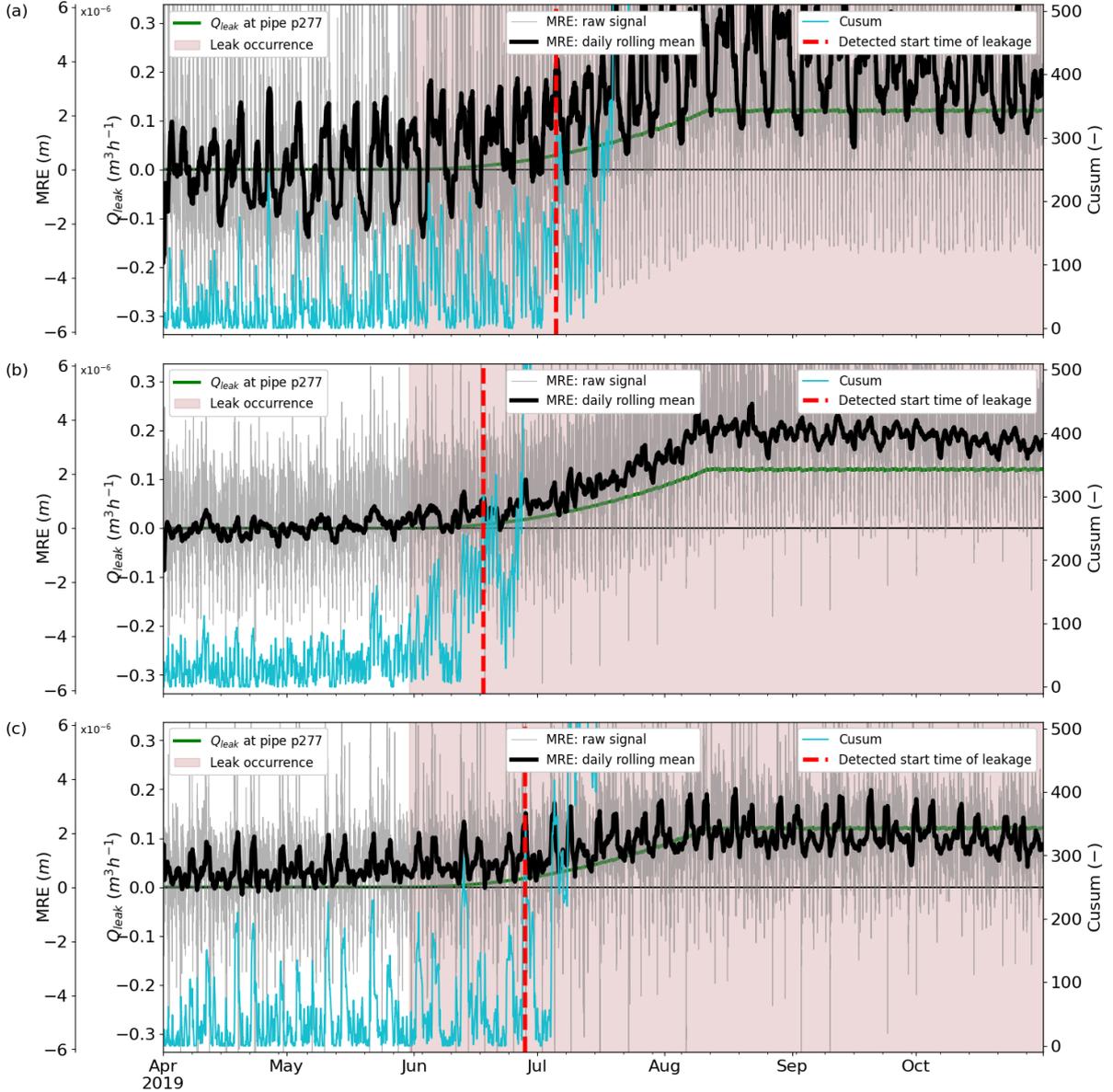

**Figure 5: Comparison of detection performance for the incipient leak (no. 2) for different algorithm versions: (a) LILA-BASE, (b) LILA-FK, (c) LILA-PINN.** MRE is plotted against the ground truth leak flow for an incipient leak starting on May 30$^{th}$, 21:55. The Cusum test statistic evaluated for the MRE raw signal is displayed on an additional axis on the right. *TTD* was calculated for $\varepsilon = 300$.

## 3.3 Uncertainty quantification

As outlined in Section 2.4, the calibration procedure of LILA-PINN is non-deterministic in contrast to both LILA-BASE and LILA-FK. To quantify this uncertainty, we trained 100 models for LILA-PINN, and analysed their leakage identification performance and *TTD*. The resulting performance metrics are reported in Table 2, while the corresponding distribution of the *TTD* is visually depicted in Figure 6.

Table 2: Leakage identification performance metrics for 100 model calibration runs.

| Leak | TP | FP | FN | *Precision* | *Recall* | *F*1-score |
|---|---|---|---|---|---|---|
| Abrupt leak | 92 | 7 | 1 | 0.93 | 0.99 | 0.96 |
| Incipient leak | 96 | 3 | 1 | 0.97 | 0.99 | 0.98 |

In the event of an abrupt leak occurrence, a total of 92 models correctly detected the leak, while seven models produced false alarms. Only one model failed to detect the leak altogether. Regarding the incipient leak scenario, 96 models correctly identified the leak, with a mere three instances of false alarms and one instance of a missed detection. Nevertheless, across both scenarios, the achieved *F*1-score surpassing 0.96 provides evidence of an overall good leak identification capability.

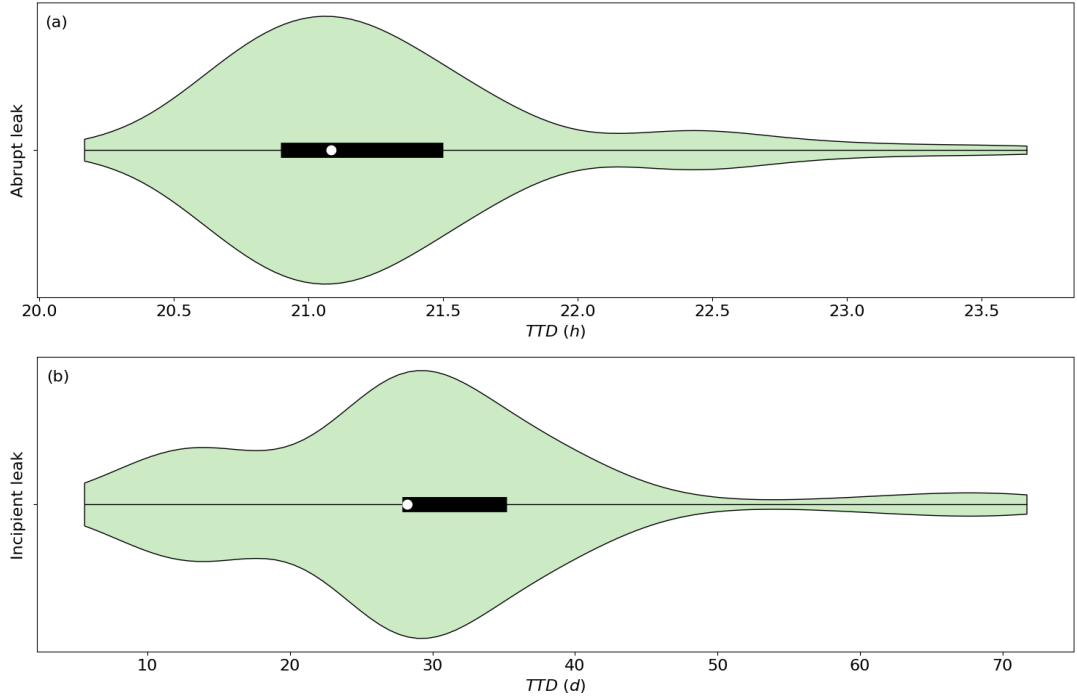

Figure 6: Uncertainty quantification of LILA-PINN for 100 model calibration runs with regards to TTD for (a) the abrupt leak and (b) the incipient leak.

For the illustration in Figure 6, only correct detections, i.e., TP, are considered for the calculation of the *TTD* for either leak. The *TTD* for the abrupt leak averages around 21.3 hours with a range from 20.2 hours to 23.7 hours. In case of the incipient leak, the average *TTD* is 29.3 days, ranging from 5.6 days to 71.7 days. While the upper range appears quite large, 75% of the LILA-PINN models still outperform LILA-BASE with a *TTD* lower than 35.6 days.

## 3.4 Sensitivity analysis

A fundamental aspect of the capacity of LILA-PINN to accurately identify leakages lies in the effectiveness of the *cusum* algorithm, which might change depending on the values of its hyperparameters consisting of the slack value $\delta$ applied during calculation of the *cusum* test statistic and the threshold value $\varepsilon$ subsequently employed to trigger the alarm. Here we complement the above leak identification results by quantifying the sensitivity of LILA-PINN to the *cusum* hyperparameters. Specifically, we test the sensitivity of all 100 LILA-PINN models for both leakages to varying *cusum* hyperparameters. The results of this sensitivity analysis are displayed in Figure 7.

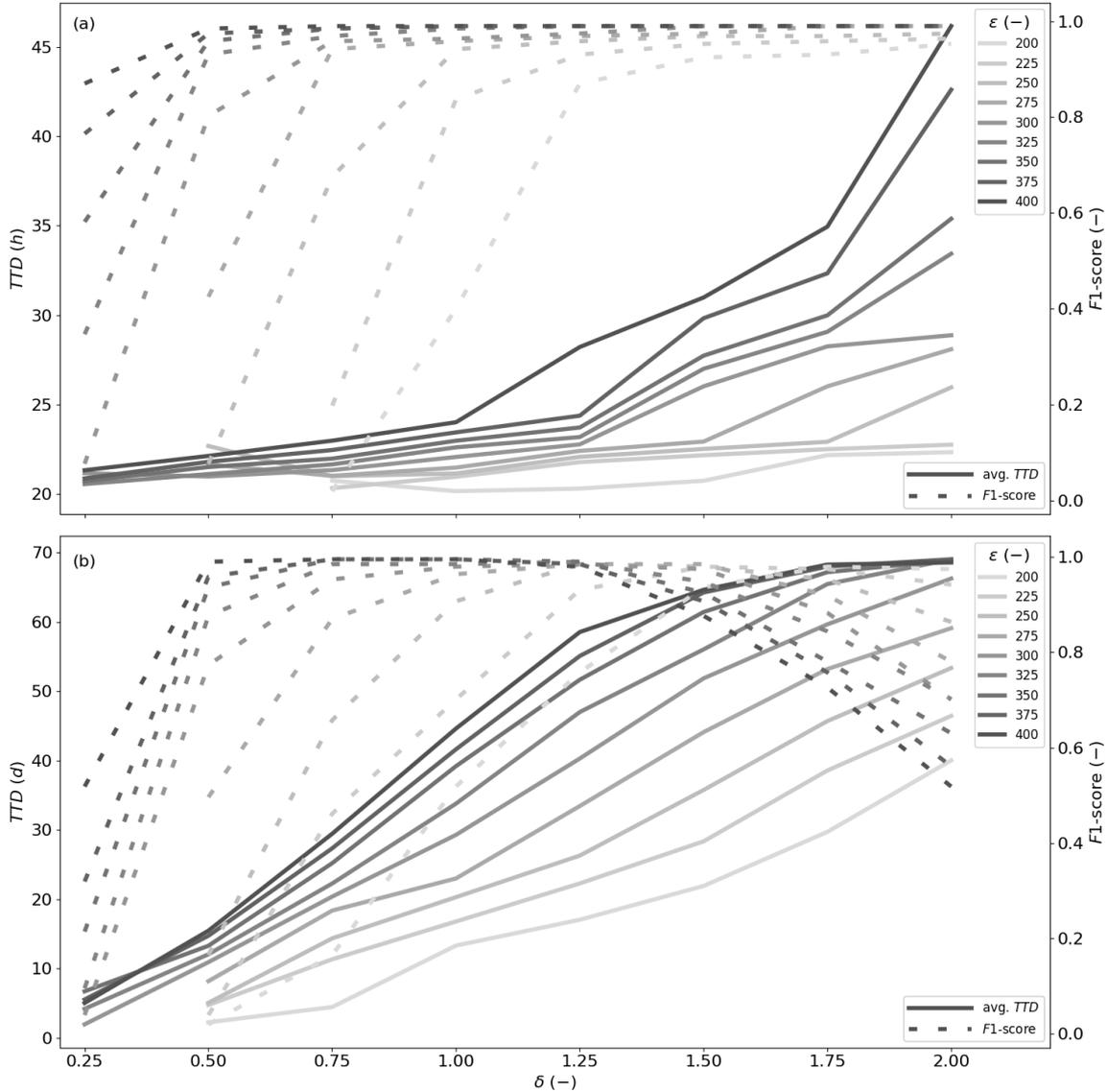

**Figure 7: Sensitivity analysis of cusum hyperparameters $\delta$ and $\varepsilon$ with regards to the average *TTD* and *F*1-score across 100 model calibration runs for (a) the abrupt leak and (b) the incipient leak.** *TTD* (solid lines) is displayed in hours for the abrupt leak and in days for the incipient leak on the left y-axis and the *F*1-score (dashed lines) is displayed on the right y-axis.

As visible in Figure 7 (a) in case of the abrupt leak, both the $F1$-score and the $TTD$ decline with a decreasing slack value $\delta$ and decreasing $\varepsilon$, which generate less restrictive conditions for raising an alarm and, thus, cause an increasing number of false alarms. However, while the $F1$-score aims to be maximized lower $TTD$ are evidently more desirable, thus, leading to a trade-off while trying to optimise both objectives simultaneously. A similar situation arises in case of the incipient leak for which the results of the sensitivity analysis are displayed in Figure 7 (b). Differently from the previous case of the abrupt leak, the $F1$-score decreases for higher slack values $\delta$ at different levels for each $\varepsilon$ due to an increased number of missed detections. Altogether, a trade-off regarding the selection of optimal values for $\delta$ and $\varepsilon$ arises when attempting to both minimise the $TTD$ and maximise the $F1$-score. A collection of Pareto-optimal (PO) solutions, along with selected suboptimal solutions, arranged in descending order based on the attained $F1$-score, are reported for both leakage cases in Tables 3 and 4, respectively.

**Table 3: Pareto-optimal (*) and selected solutions for $\delta$ and $\varepsilon$ with regards to $TTD$ and $F1$-score for the abrupt leak.**

| $\delta$ | $\varepsilon$ | PO | avg. $TTD$ | $F1$-score |
|---|---|---|---|---|
| (-) | (-) | | (h) | (-) |
| 1.00 | 350 | (*) | 23.0 | 0.990 |
| 0.75 | 350 | (*) | 22.0 | 0.985 |
| 0.50 | 400 | | 22.1 | 0.985 |
| 0.50 | 375 | (*) | 21.8 | 0.974 |
| 0.75 | 325 | (*) | 21.7 | 0.969 |

**Table 4: Pareto-optimal (*) and selected solutions for $\delta$ and $\varepsilon$ with regards to $TTD$ and $F1$-score for the incipient leak.**

| $\delta$ | $\varepsilon$ | PO | avg. $TTD$ | $F1$-score |
|---|---|---|---|---|
| (-) | (-) | | (d) | (-) |
| 0.75 | 350 | (*) | 25.2 | 0.995 |
| 0.50 | 400 | (*) | 15.4 | 0.990 |
| 0.75 | 325 | | 22.3 | 0.985 |
| 1.50 | 225 | | 28.3 | 0.980 |

The results in Table 3 reveal a distinct trade-off where a simultaneous reduction in $TTD$ and $F1$-score is observed for the abrupt leakage scenario. Conversely, as reported in Table 4, the incipient leak scenario showcases an initial decline in $TTD$, reaching an optimal value of 15.4 days, while maintaining an excellent leakage identification performance with an $F$1-score of 0.99. Furthermore, subsequent decreases in $F$1-score result in a significant increase in $TTD$, indicating a departure from Pareto optimality.

Ultimately, a favourable trade-off is found for values of $\delta=0.5$ and $\varepsilon=400$, resulting in the maintenance of high $F$1-scores of 0.985 and 0.99 for the abrupt and incipient leak scenarios, respectively, while achieving excellent $TTD$ of 22.1 hours and 15.4 days. In comparison to the performance of the baseline algorithm, LILA-BASE, the $TTD$ has exhibited a substantial reduction by a factor of 5.4, decreasing from 119 hours to 22.1 hours for the abrupt leak, and a factor of 2.3, decreasing from 35.6 days to 15.4 days for the incipient leak. Moreover, compared to the results from the original work in Daniel et al. (2022), $TTD$ could be reduced by a factor of 5.3 from 116 hours for the abrupt leak and a factor of 3.0 from 45.9 days for the incipient leak.

## 4  Discussion

While these results provide evidence of a significant improvement on leakage identification, the physics-informed ML approach also entails several limitations worth mentioning.

First, due to the only aggregated observability of the irregular demands, the problem of extracting them from the pressure data remains underdefined. This introduces a degree of uncertainty into the algorithm's performance affecting not only the time to detection but also the identification ability, leading to an increased probability for false alarms and missed events. Nonetheless, these issues can be mitigated by appropriately adjusting the hyperparameters of the *cusum* algorithm facilitating change point detection. By optimizing these parameters, it is possible to achieve a satisfactory containment of the loss in identification ability while significantly improving the detection time.

On the other hand, the process of tuning these hyperparameters poses a challenge as it currently relies on semi-manual adjustment, which necessitates additional knowledge about the presence of previous leaks. Alternatively, utility companies may perform fire-flow tests to simulate leakages and acquire optimal hyperparameters. However, it is worth noting that *cusum* hyperparameters are sensitive to both noise and irregular demands. While a comprehensive sensitivity analysis is conducted in the scope of this study, a more generalized approach to their selection would be imperative for widespread practical application.

Lastly, in this study, we omitted the inclusion of the leakage localisation task with the goal of ultimately providing an entirely data-based leakage identification and localisation algorithm, thereby completely surpassing the restriction of model-based algorithms for a well-calibrated hydraulic model. Within the context of this study, the results from the original work regarding the most affected sensor remain valid with regards to leakage localisation (Daniel et al., 2022). While this information already greatly reduces the search radius for ground personnel, other leakage localisation algorithms that do incorporate information from hydraulic models are capable of pinpointing leakages to the level of individual pipes. Therefore, to practically implement our proposed data-based approach, a higher spatial resolution for the localisation of identified leaks would be required than currently possible with this version of our algorithm.

## 5   Conclusions

In this work, we presented a physics-informed machine learning model for leakage identification in water distribution networks that can extract and further incorporate irregular water demands, thus, increasing robustness against missed detections. The model utilises sub-hourly data from pressure sensors and identifies leakages, once calibrated on leakage-free data, by subjecting the model reconstruction error to further change point detection analysis employing the *cusum* algorithm. The machine learning model applied in this work is based on a fully connected neural network which estimates the irregular demands implicitly contained in the pressure data by establishing a feedback loop with a linear regression layer. The physical information of the hydraulic network reflecting the Bernoulli principle is thus incorporated as an inductive bias within this regression layer.

The advantages of incorporating physics-informed machine learning are twofold. Firstly, by explicitly predicting the unknown irregular demands from the observable pressure data by means of a neural network, additional knowledge about their temporal trajectories is acquired. Secondly, feeding these predictions back into the leakage identification algorithm enhances its robustness by removing irregular demand patterns from the data presented to the change point detection algorithm, hence, reducing noise and speeding up leakage identification.

To evaluate our algorithm, we apply it to the benchmark leakage dataset from the L-town water distribution network created for the BattLeDIM (Vrachimis et al., 2022). Specifically, we consider the sensor data obtained from three pressure sensors in DMA C of the L-town network, where three industrial demands that classify as irregular demands were imposed on the hydraulic network, together with two leakages, one abrupt and one incipient.

By applying our algorithm, we find that the neural network tasked with estimating the irregular demands is capable of accurately extracting two out of the three industrial demand patterns located in DMA C of L-town, accurately capturing their temporal trajectories and peak magnitudes. However, instead of predicting the third industrial demand, the model identifies residuals of diurnal demand patterns, which are ultimately found to be more relevant for the leakage identification task. Feeding back these estimates into the linear regression model, the residual noise in the model reconstruction error is significantly reduced, leading to improved leakage identification. In comparison to the original model, the time to detection decreases by a factor of up to 5.3 to only 22.1 hours and by a factor of 3.0 to only 15.4 days for the abrupt and incipient leak in DMA C, respectively.

Overall, contributions and limitations of this work reveal several avenues for further research. First, manual selection of *cusum* hyperparameters remains an issue that is only partially tackled with the sensitivity analysis conducted within the context of this study. Further investigation may specifically focus on the relation between these hyperparameters and the level of noise present in the sensor data as well as the relative magnitude of present irregular demands compared to prominent diurnal patterns. Second, data-based leakage localisation needs to be improved for the ability to compete with model-based approaches. Here, further analysis of the model reconstruction error under consideration of geospatial data may yield improved localisation information and aid overall leakage localisation accuracy.

## Declaration of competing interests

The authors declare no competing interests.

## Data and Code availability

All models and code generated or used during this study are available in online repositories (Daniel and Cominola, 2023). All data used during the study are available in online repositories (Vrachimis et al., 2020) in accordance with funder data retention policies.

## Acknowledgements

The authors would like to thank the BattLeDIM committee for organising the competition and the creation of the benchmark dataset used in this study. Moreover, the authors acknowledge the support of the Helmholtz Einstein International Berlin Research School in Data Science (HEIBRiDS).

## References

Arregui, F., Cobacho, R., Soriano, J., Jimenez-Redal, R., 2018. Calculation Proposal for the Economic Level of Apparent Losses (ELAL) in a Water Supply System. Water 10, 1809. https://doi.org/10.3390/w10121809

Barton, N.A., Farewell, T.S., Hallett, S.H., Acland, T.F., 2019. Improving pipe failure predictions: Factors affecting pipe failure in drinking water networks. Water Research 164, 114926. https://doi.org/10.1016/j.watres.2019.114926

Brunton, S.L., Kutz, J.N., 2019. Data-Driven Science and Engineering: Machine Learning, Dynamical Systems, and Control, 1st ed. Cambridge University Press. https://doi.org/10.1017/9781108380690

Daniel, I., Cominola, A., 2023. LILA-PINN (Code) [WWW Document]. URL https://github.com/SWN-group-at-TU-Berlin/LILA-PINN (accessed 7.14.23).

Daniel, I., Pesantez, J., Letzgus, S., Khaksar Fasaee, M.A., Alghamdi, F., Berglund, E., Mahinthakumar, G., Cominola, A., 2022. A Sequential Pressure-Based Algorithm for Data-Driven Leakage Identification and Model-Based Localization in Water Distribution Networks. J. Water Resour. Plann. Manage. 148, 04022025. https://doi.org/10.1061/(ASCE)WR.1943-5452.0001535

Eliades, D.G., Polycarpou, M.M., 2012. Leakage fault detection in district metered areas of water distribution systems. Journal of Hydroinformatics 14, 992–1005. https://doi.org/10.2166/hydro.2012.109

Flörke, M., Schneider, C., McDonald, R.I., 2018. Water competition between cities and agriculture driven by climate change and urban growth. Nat Sustain 1, 51–58. https://doi.org/10.1038/s41893-017-0006-8

Fu, G., Jin, Y., Sun, S., Yuan, Z., Butler, D., 2022. The role of deep learning in urban water management: A critical review. Water Research 223, 118973. https://doi.org/10.1016/j.watres.2022.118973

Goodfellow, I., Bengio, Y., Courville, A., 2016. Deep learning, Adaptive computation and machine learning. The MIT Press, Cambridge, Massachusetts.

Guo, G., Yu, X., Liu, S., Ma, Z., Wu, Y., Xu, X., Wang, X., Smith, K., Wu, X., 2021. Leakage Detection in Water Distribution Systems Based on Time–Frequency Convolutional Neural Network. J. Water Resour. Plann. Manage. 147, 04020101. https://doi.org/10.1061/(ASCE)WR.1943-5452.0001317

Hutton, C., Kapelan, Z., 2015. Real-time Burst Detection in Water Distribution Systems Using a Bayesian Demand Forecasting Methodology. Procedia Engineering 119, 13–18. https://doi.org/10.1016/j.proeng.2015.08.847

Kammoun, M., Kammoun, A., Abid, M., 2022. Leak Detection Methods in Water Distribution Networks: A Comparative Survey on Artificial Intelligence Applications. J. Pipeline Syst. Eng. Pract. 13, 04022024. https://doi.org/10.1061/(ASCE)PS.1949-1204.0000646

Karniadakis, G.E., Kevrekidis, I.G., Lu, L., Perdikaris, P., Wang, S., Yang, L., 2021. Physics-informed machine learning. Nat Rev Phys 3, 422–440. https://doi.org/10.1038/s42254-021-00314-5

Klise, K.A., Bynum, M., Moriarty, D., Murray, R., 2017. A software framework for assessing the resilience of drinking water systems to disasters with an example earthquake case study. Environmental Modelling & Software 95, 420–431. https://doi.org/10.1016/j.envsoft.2017.06.022

Konapala, G., Mishra, A.K., Wada, Y., Mann, M.E., 2020. Climate change will affect global water availability through compounding changes in seasonal precipitation and evaporation. Nat Commun 11, 3044. https://doi.org/10.1038/s41467-020-16757-w

Lee, P.J., Vítkovský, J.P., Lambert, M.F., Simpson, A.R., Liggett, J.A., 2005. Frequency Domain Analysis for Detecting Pipeline Leaks. J. Hydraul. Eng. 131, 596–604. https://doi.org/10.1061/(ASCE)0733-9429(2005)131:7(596)

Lever, J., Krzywinski, M., Altman, N., 2016. Classification evaluation. Nat Methods 13, 603–604. https://doi.org/10.1038/nmeth.3945

Levinas, D., Perelman, G., Ostfeld, A., 2021. Water Leak Localization Using High-Resolution Pressure Sensors. Water 13, 591. https://doi.org/10.3390/w13050591

Liemberger, R., Wyatt, A., 2019. Quantifying the global non-revenue water problem. Water Supply 19, 831–837. https://doi.org/10.2166/ws.2018.129

Liu, Y., Hejazi, M., Kyle, P., Kim, S.H., Davies, E., Miralles, D.G., Teuling, A.J., He, Y., Niyogi, D., 2016. Global and Regional Evaluation of Energy for Water. Environ. Sci. Technol. 50, 9736–9745. https://doi.org/10.1021/acs.est.6b01065

Maas, A.L., Hannun, A.Y., Ng, A.Y., 2013. Rectifier Nonlinearities Improve Neural Network Acoustic Models, in: Proceedings of the 30th International Conference on Machine Learning. JMLR.org, Atlanta, Georgia, USA, p. 6.

Mansour-Rezaei, S., Naser, G., 2013. Contaminant intrusion in water distribution systems: An ingress model. Journal - American Water Works Association 105, E29–E39. https://doi.org/10.5942/jawwa.2013.105.0001

Misiunas, D., Vítkovský, J., Olsson, G., Lambert, M., Simpson, A., 2006. Failure monitoring in water distribution networks. Water Science and Technology 53, 503–511. https://doi.org/10.2166/wst.2006.154

Montgomery, D.C., 2020. Introduction to statistical quality control, Eighth edition. ed. John Wiley & Sons, Inc, Hoboken, NJ.

Palau, C.V., Arregui, F.J., Carlos, M., 2012. Burst Detection in Water Networks Using Principal Component Analysis. Journal of Water Resources Planning and Management 138, 47–54. https://doi.org/10.1061/(ASCE)WR.1943-5452.0000147

Palmitessa, R., Grum, M., Engsig-Karup, A.P., Löwe, R., 2022. Accelerating hydrodynamic simulations of urban drainage systems with physics-guided machine learning. Water Research 223, 118972. https://doi.org/10.1016/j.watres.2022.118972

Parkinson, S., 2021. Guiding urban water management towards 1.5 °C. npj Clean Water 4, 34. https://doi.org/10.1038/s41545-021-00126-1

Paszke, A., Gross, S., Massa, F., Lerer, A., Bradbury, J., Chanan, G., Killeen, T., Lin, Z., Gimelshein, N., Antiga, L., Desmaison, A., Kopf, A., Yang, E., DeVito, Z., Raison, M., Tejani, A., Chilamkurthy, S., Steiner, B., Fang, L., Bai, J., Chintala, S., 2019. PyTorch: An Imperative Style, High-Performance Deep Learning Library. Presented at the 33rd Conference on Neural Information Processing Systems (NeurIPS 2019), Vancouver, Canada, p. 12.

Puust, R., Kapelan, Z., Savic, D.A., Koppel, T., 2010. A review of methods for leakage management in pipe networks. Urban Water Journal 7, 25–45. https://doi.org/10.1080/15730621003610878

Rajabi, M.M., Komeilian, P., Wan, X., Farmani, R., 2023. Leak detection and localization in water distribution networks using conditional deep convolutional generative adversarial networks. Water Research 238, 120012. https://doi.org/10.1016/j.watres.2023.120012

Ramsundar, B., Zadeh, R.B., 2018. TensorFlow for deep learning: from linear regression to reinforcement learning, First edition. ed. O'Reilly Media, Beijing.

Romero-Ben, L., Alves, D., Blesa, J., Cembrano, G., Puig, V., Duviella, E., 2023. Leak detection and localization in water distribution networks: Review and perspective. Annual Reviews in Control 55, 392–419. https://doi.org/10.1016/j.arcontrol.2023.03.012

Shortridge, J.E., Guikema, S.D., 2014. Public health and pipe breaks in water distribution systems: Analysis with internet search volume as a proxy. Water Research 53, 26–34. https://doi.org/10.1016/j.watres.2014.01.013

Soldevila, A., Fernandez-Canti, R.M., Blesa, J., Tornil-Sin, S., Puig, V., 2016. Leak localization in water distribution networks using model-based Bayesian reasoning, in: 2016 European Control Conference (ECC). Presented at the 2016 European Control
22

Conference (ECC), IEEE, Aalborg, Denmark, pp. 1758–1763. https://doi.org/10.1109/ECC.2016.7810545

Steffelbauer, D.B., Deuerlein, J., Gilbert, D., Abraham, E., Piller, O., 2022. Pressure-Leak Duality for Leak Detection and Localization in Water Distribution Systems. J. Water Resour. Plann. Manage. 148, 04021106. https://doi.org/10.1061/(ASCE)WR.1943-5452.0001515

Taormina, R., Galelli, S., Tippenhauer, N.O., Salomons, E., Ostfeld, A., Eliades, D.G., Aghashahi, M., Sundararajan, R., Pourahmadi, M., Banks, M.K., Brentan, B.M., Campbell, E., Lima, G., Manzi, D., Ayala-Cabrera, D., Herrera, M., Montalvo, I., Izquierdo, J., Luvizotto, E., Chandy, S.E., Rasekh, A., Barker, Z.A., Campbell, B., Shafiee, M.E., Giacomoni, M., Gatsis, N., Taha, A., Abokifa, A.A., Haddad, K., Lo, C.S., Biswas, P., Pasha, M.F.K., Kc, B., Somasundaram, S.L., Housh, M., Ohar, Z., 2018. Battle of the Attack Detection Algorithms: Disclosing Cyber Attacks on Water Distribution Networks. J. Water Resour. Plann. Manage. 144, 04018048. https://doi.org/10.1061/(ASCE)WR.1943-5452.0000969

Tornyeviadzi, H.M., Seidu, R., 2023. Leakage detection in water distribution networks via 1D CNN deep autoencoder for multivariate SCADA data. Engineering Applications of Artificial Intelligence 122, 106062. https://doi.org/10.1016/j.engappai.2023.106062

Vadyala, S.R., Betgeri, S.N., Matthews, J.C., Matthews, E., 2022. A review of physics-based machine learning in civil engineering. Results in Engineering 13, 100316. https://doi.org/10.1016/j.rineng.2021.100316

Vrachimis, S.G., Eliades, D.G., Taormina, R., Kapelan, Z., Ostfeld, A., Liu, S., Kyriakou, M., Pavlou, P., Qiu, M., Polycarpou, M.M., 2022. Battle of the Leakage Detection and Isolation Methods. J. Water Resour. Plann. Manage. 148, 04022068. https://doi.org/10.1061/(ASCE)WR.1943-5452.0001601

Vrachimis, S.G., Eliades, D.G., Taormina, R., Ostfeld, A., Kapelan, Z., Liu, S., Kyriakou, M.S., Pavlou, P., Qiu, M., Polycarpou, M., 2020. Dataset of BattLeDIM: Battle of the Leakage Detection and Isolation Methods. https://doi.org/10.5281/ZENODO.4017659

Wan, X., Kuhanestani, P.K., Farmani, R., Keedwell, E., 2022. Literature Review of Data Analytics for Leak Detection in Water Distribution Networks: A Focus on Pressure and Flow Smart Sensors. J. Water Resour. Plann. Manage. 148, 03122002. https://doi.org/10.1061/(ASCE)WR.1943-5452.0001597

Zaman, D., Tiwari, M.K., Gupta, A.K., Sen, D., 2020. A review of leakage detection strategies for pressurised pipeline in steady-state. Engineering Failure Analysis 109, 104264. https://doi.org/10.1016/j.engfailanal.2019.104264

Zhou, X., 2019. Deep learning identifies accurate burst locations in water distribution networks. Water Research 12.